\documentclass{article}

\usepackage[final]{corl_2021} % Uncomment for the camera-ready ``final'' version.

\usepackage[numbers]{natbib}
\usepackage{multicol}
\hypersetup{
    colorlinks=true,
    linkcolor=orange,
    filecolor=magenta,      
    urlcolor=orange,
    citecolor=orange,
}

\usepackage{pdfx}

\usepackage{wrapfig}
\usepackage{graphicx}
\usepackage{xcolor}
\usepackage{amssymb,amsthm}
\usepackage{physics}
\usepackage{algorithm}
\usepackage{algpseudocode}
\usepackage{comment}
\usepackage{cleveref}
\usepackage{hhline}
\usepackage{booktabs}
\usepackage{wrapfig}
\usepackage{balance}
\usepackage[small]{caption}
\usepackage{enumitem}
\usepackage{multirow}
\usepackage{colortbl}

\setlist{leftmargin=5.5mm}
\usepackage{titlesec}
\titlespacing*{\subsection}{0pt}{0.002\baselineskip}{0.001\baselineskip}
\titlespacing*{\section}{0pt}{0.01\baselineskip}{0.005\baselineskip}
\usepackage{thm-restate}
\newif\ifrevision

\Crefname{equation}{Eq.}{Eqs.}
\Crefname{figure}{Fig.}{Figs.}
\def\argq#1{\underset Q{\arg#1}\;}
\newcommand\params[1][|]{\param#1\end}
\def\param#1|#2\end{\qty{#1\alpha_m#2, #1\omega_m#2}_{m=1}^M}
\def\D{\mathcal D}
\def\H{\mathcal H}

\let\cref=\Cref

\renewcommand\paragraph{\noindent\textbf}

\makeatletter
\let\@figure@\figure
\let\@wrapfigure@\wrapfigure
\let\@table@\table
\def\csetup{\captionsetup{labelfont={color=\col},font={small,color=\col}}}
\newenvironment{revision}{%
\ifrevision%
\def\col{blue}%
\color{\col}%
\csetup%
\renewcommand\figure[1][]{\@figure@[##1]%
\csetup\color\col}%
\renewcommand\wrapfigure[3][]{\@wrapfigure@[##1]{##2}{##3}%
\csetup\color\col}%
\renewcommand\table[1][]{\@table@[##1]%
\captionsetup{labelfont={color=\col},font={color=\col}}\color\col%
}%
\fi{}%
}{\ifrevision\hskip -3.8pt\fi}
\makeatother

\newcommand\compactdots{\hbox to 0.8em{.\hss.\hss.}}

\makeatletter
\renewcommand\table[1][]{\@table@[#1]\captionsetup{font={normal}}}

\makeatother
\title{Learning Multimodal Rewards from Rankings}

\author{
  Vivek Myers $^\dagger$\qquad Erdem B\i y\i k $^\ddagger$\qquad Nima Anari $^\dagger$\qquad Dorsa Sadigh $^{\dagger,\ddagger}$\\
  $^\dagger$ Department of Computer Science, Stanford University\\
  $^\ddagger$ Department of Electrical Engineering, Stanford University\\
  \texttt{\{vmyers,ebiyik\}@stanford.edu, \{anari,dorsa\}@cs.stanford.edu}
}

\begin{document}
\maketitle

\abovedisplayskip=0pt
\abovedisplayshortskip=0pt
\belowdisplayskip=0pt
\belowdisplayshortskip=0pt

%===============================================================================
\vspace{-2em}
\begin{abstract}
    Learning from human feedback has shown to be a useful approach in acquiring robot reward functions. However, expert feedback is often assumed to be drawn from an underlying \emph{unimodal} reward function. This assumption does not always hold including in settings where multiple experts provide data or when a single expert provides data for different tasks---we thus go beyond learning a unimodal reward and focus on learning a \emph{multimodal} reward function. We formulate the multimodal reward learning as a mixture learning problem and develop a novel \emph{ranking}-based learning approach, where the experts are only required to rank a given set of trajectories. Furthermore, as access to interaction data is often expensive in robotics, we develop an \emph{active} querying approach to accelerate the learning process. We conduct experiments and user studies using a multi-task variant of OpenAI's LunarLander and a real Fetch robot, where we collect data from multiple users with different preferences. The results suggest that our approach can efficiently learn multimodal reward functions, and improve data-efficiency over benchmark methods that we adapt to our learning problem.
\end{abstract}


%===============================================================================

\section{Introduction}

\begin{wrapfigure}{R}{.35\columnwidth}
    \centering
    \vspace*{-10px}
    \includegraphics[width=\linewidth]{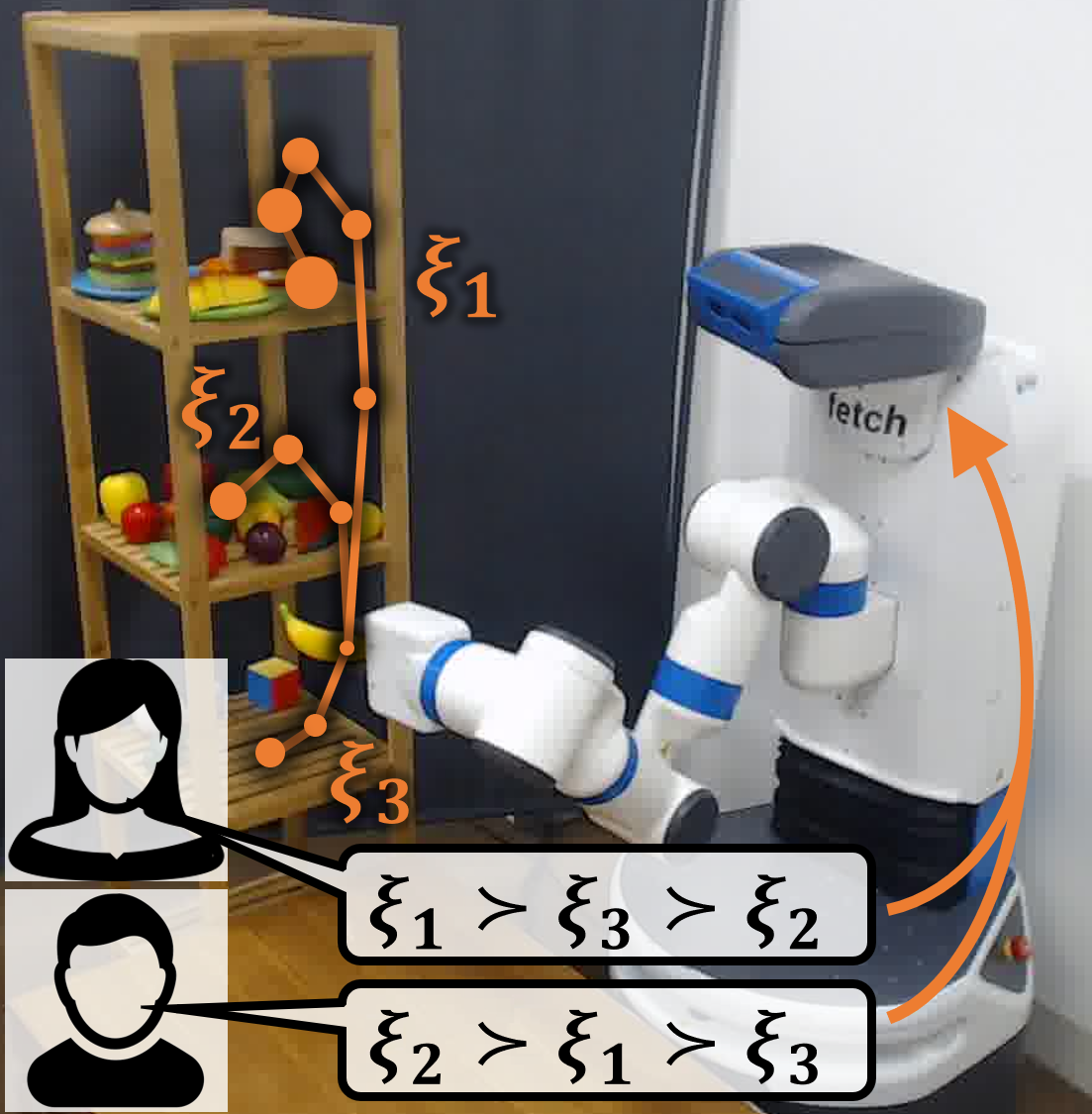}
    \vspace*{-15px}
    \caption{Fetch robot putting a banana on one of the three shelves. The two users have different preferences, and so they provide different rankings to the robot. The robot needs to be able to model multimodal reward functions for successfully achieving the task.}
    \vspace*{-15px}
    \label{fig:robot_experiment_visual}
\end{wrapfigure}

Learning a reward function from different sources of human data is a fundamental problem in robot learning.
In recent years, there has been a large body of work that learns reward functions using various forms of human input, such as expert demonstrations \cite{ziebart2008maximum}, suboptimal demonstrations~\cite{cao2021learning,grollman2011donut,wu2019imitation}, pairwise comparisons \cite{biyik2020learning,biyik2020active}, physical corrections \cite{bajcsy2018learning,li2021learning}, rankings \cite{brown2019extrapolating}, and trajectory assessments \cite{shah2020interactive}. These works focus on learning a \emph{unimodal} reward function that models human preferences on a target task. However, this unimodality assumption does not always hold: human preferences are usually more complex and need to be captured via a multimodal representation. Further, even if the preferences of a human are truly unimodal, we often use a mixture of data from multiple humans, which can be difficult to disentangle, leading to multimodality.
% formulating the reward design problem in this way brings an important limitation: the underlying reward function has to be \emph{unimodal}.

As an example, consider a robot placing a banana on one of the three shelves (see Fig.~\ref{fig:robot_experiment_visual}). The middle shelf is often used for fruits, but it has no room left and if the robot tries to put the banana there, it may cause other fruits to fall. The top shelf has some space but it has been used for cooked meals. The bottom shelf has a lot of free space, but is usually used only for toys. In such a scenario, people may have very different preferences about what the robot should do. If we try to learn a unimodal reward using data collected from multiple people, the robot is likely to fail in the task, because the data will include inconsistent preferences.

%As another example, consider the driving scenario presented in \cite{cao2020reinforcement} (see Fig.~\ref{fig:driving_visual}), where the red car attempts to make an unprotected left turn, but fails to observe the blue car occluded behind a truck. An aggressive driver might accelerate and avoid the accident by completing the turn before the blue car reaches the collision point. Similarly, a timid driver would move slowly and brake sharply the moment it sees the blue car, which also prevents the accident. Even though both modes can avoid the accident, a driving policy learned by a mixture of this data is likely to fail while trying to comply with both modes. In fact, \citet{cao2020reinforcement} demonstrate simply applying imitation learning using this data fails in this case, and a separation of different modes is needed.

One solution is of course to label the different modes in the data. For example, one could separate the data based on the preferred shelf, and learn different reward functions for each shelf. However, this separation is not always straightforward. For example in a driving dataset, it is unclear what should be labeled as aggressive or timid driving. Clustering the data based on the human who provided the data is also not viable as it will introduce data-inefficiency issues, and perhaps more importantly, humans are not always unimodal: a usually timid driver can drive more aggressively when in a hurry.

These examples motivate us to develop methods that can learn \emph{multimodal} reward functions using datasets that are not specifically labeled with the modes. To this end, previous work proposed learning from demonstrations to learn multimodal policies \cite{hausman2018multi,fei2020triple} or reward functions with multiple possible intentions \cite{babes2011apprenticeship,ramponi2020truly}. However, learning from expert demonstrations is often extremely challenging in robotics as providing demonstrations on a robot with high degrees of freedom is non-trivial \cite{biyik2020learning,akgun2012keyframe}, and humans have difficulty giving demonstrations that align with their preferences due to their cognitive biases \cite{basu2017you,kwon2020when}. Thus, it is desirable to have methods that learn from other more reliable sources of data.
For instance, humans can reliably compare two different trajectories, enabling a robot to learn from pairwise comparisons
% such as pairwise comparisons as humans can reliably compare different trajectories 
\cite{biyik2020learning,brown2019extrapolating,basu2018learning}.
%However, existing works on learning multimodal rewards from expert demonstrations rely on imitation learning techniques,
% such as GAIL \cite{ho2016generative}, 
%and do not handle other data sources such as pairwise comparisons.

%\begin{wrapfigure}{R}{.35\columnwidth}
%    \centering
%    \vspace*{-15px}
%    \includegraphics[width=\linewidth]{figs/driving_visual.png}
%    \vspace*{-15px}
%    \caption{The red car does not observe the blue car due to the occluding truck until it comes to the intersection. It is possible to avoid accident by \textbf{(left)} completing the turn aggressively or \textbf{(right)} making a hard-brake. An autonomous vehicle trying to learn from such mixture data must be able to model multimodal reward functions for safe and efficient driving.}
%    \vspace*{-15px}
%    \label{fig:driving_visual}
%\end{wrapfigure}

While learning from pairwise comparisons provides a rich source of data for learning reward functions, the theoretical results by \citet{zhao2016learning} imply that extending the existing comparison-based reward learning techniques to multimodal reward functions is not possible, i.e. failure cases can be constructed, where pairwise comparisons are not sufficient for identifying different modes of a multimodal function. Our insight is that it is possible to learn a multimodal reward function by going beyond pairwise comparisons and instead using \emph{rankings}.

%We want to emphasize this is different from learning nonlinear rewards. Nonlinear rewards allow users to have multiple sets of desired behavior: a user may prefer both the top and the bottom shelves over the middle shelf. However, the existing methods that handle nonlinearities assume unimodal human feedback: the user must have a consistent preference towards either of the modes. Thus, either (\emph{top} $\succ$ \emph{bottom} $\succ$ \emph{middle}) or (\emph{bottom} $\succ$ \emph{top} $\succ$ \emph{middle}) is expected. In this work, we relax this assumption and learn multimodal rewards without requiring consistent rankings in the dataset. Our framework allows both (\emph{top} $\succ$ \emph{bottom} $\succ$ \emph{middle}) and (\emph{bottom} $\succ$ \emph{top} $\succ$ \emph{middle}) to be in the dataset, and we recover both modes.

To achieve this, we formulate multimodal reward learning as a mixture learning problem. As data is a very expensive resource in robotics, we further develop an \emph{active querying} method that aims to ask the human users the most informative queries. Our contributions are three-fold:
\begin{itemize}[nosep]
    \item We develop a method that uses rankings from humans to learn multimodal reward functions.
    \item We develop an active querying method to improve data-efficiency by asking the most informative ranking queries.
    \item We conduct extensive experiments and user studies with OpenAI's LunarLander and a real Fetch robot to test our learning and querying methods in comparison to baselines.
\end{itemize}

\section{Related Work}
%In this section, we will go over the existing literature on reward learning and preference-based learning in robotics. We will then discuss the recent theoretical works on learning mixture models from rankings which inform our proposed approach.

\noindent\textbf{Reward Learning in Robotics.} Learning reward functions from human feedback is a fundamental problem in robot learning. \citet{ng2000algorithms} and \citet{abbeel2004apprenticeship} introduce the problem of learning from demonstrations in the space of robotics. Later works focus on improving \emph{inverse reinforcement learning} algorithms by reducing the ambiguity in learned rewards \cite{ziebart2008maximum,ratliff2006maximum}.

Due to the difficulty in providing expert demonstrations in robotics, recent works attempt to learn reward functions using other forms of human feedback, such as physical corrections \cite{bajcsy2018learning,li2021learning} and pairwise comparisons where a human user compares the quality of two robot trajectories \cite{cakmak2011human,akrour2012april,sadigh2017active,christiano2017deep}. Later works extend these algorithms for better time and data-efficiency \cite{biyik2018batch,wilde2020active,biyik2019asking}. Though there have been works to extend this framework to reward functions modeled with Gaussian processes to capture nonlinearities \cite{biyik2020active,tucker2020preference,li2020roial}, the underlying reward function has always been unimodal.

\noindent\textbf{Preference-based Learning.} Outside of robotics, preference-based learning, where data are in the form of comparisons, selections out of a set, or rankings, has attracted attention due to the ease of collecting data and its reliability. Prior works have studied this in the classification \cite{chen2017near}, bandits \cite{busa2014survey}, and reinforcement learning settings \cite{wirth2017survey}.

%Learning from comparisons or rankings goes back to the algorithms that try to sort a discrete set of items based on their utilities \cite{braverman2008noisy,braverman2009sorting}.
In our problem, we have a continuous hypothesis space of reward functions. In such cases, it is common to model human comparisons or rankings with a computational model. Bradley-Terry is one such model for pairwise comparisons \cite{chen2013pairwise}, which is easily extended to the queries where the human chooses the best of multiple items. Known as multinomial logits (MNL) \cite{chen2018nearly}, this model has been widely used for human preferences in many fields \cite{ben2018discrete,wilson2019ten,biyik2019green} including robotics \cite{sadigh2017active,biyik2018batch,basu2018learning}.

To extend these models to rankings, Plackett-Luce \cite{maystre2015fast,archambeau2012plackett} and Mallows models \cite{lu2011learning,vitelli2018probabilistic,lu2014effective,busa2014preference} are commonly employed. In this paper, we use the Plackett-Luce model as it is a natural extension of MNL, which is widely used in robotics with great success. We formalize this model in Section~\ref{sec:problem}.

Even though there has been much research in this domain, all works we mentioned here focus on the unimodal case, and do not work with the multimodal preferences. %More recently, researchers have considered using mixture models to learn multimodal preferences. We next discuss these works.

\noindent\textbf{Learning Mixture Models from Rankings.}
One way to model multimodal reward functions is through mixture models, where the data is assumed to come from different individual models with some unknown probabilities. To this end, previous works consider mixtures of MNLs \cite{de2010bayesian,chierichetti2018learning}, Plackett-Luce models \cite{zhao2016learning}, and Mallows models \cite{liu2018efficiently}. Other works adopt different methods to model multimodality, such as by assuming latent state dynamics that transition between different modes \cite{morton2017simultaneous,basu2019active} or by learning the different modes from labeled datasets \cite{cao2020reinforcement,qureshi2019composing}. To avoid these modeling assumptions, we focus on directly learning the mixture model.

While \citet{zhao2016learning} have theoretically studied the mixture of Plackett-Luce choice models, which also informs our algorithm in terms of the query sizes, they only focus on learning the rewards of a discrete set of items. To the best of our knowledge, our paper is the first work that deals with a continuous hypothesis space under a mixture of Plackett-Luce models. Furthermore, we propose an active querying strategy for this mixture model to improve data-efficiency for human-in-the-loop learning, which is crucial in data-hungry applications such as robotics.

\section{Problem Formulation} 
\label{sec:problem}
\label{sec:formulation}

\noindent\textbf{Setup.}
We consider a fully-observable deterministic dynamical system. A trajectory $\xi$ in this system is a series of states and actions, i.e., $\xi=(s_0,a_0,\ldots, s_T, a_T)$. The set of feasible trajectories is $\Xi$.

We assume there is a set of $M$ individual reward functions that are possibly different, 
%For example, these might be people with different preferences (aggressive and timid drivers) or a person might be trying to teach multiple different tasks to a robot. 
each of which encodes some preference between the trajectories in $\Xi$. For the rest of the formalism, we refer to each individual reward function as an \emph{expert} for the clarity of the presentation.

Following the common linearity assumption in reward learning \cite{ziebart2008maximum,biyik2020learning,wilde2020active}, we assume each preference can be modeled as a linear reward function over a known fixed feature space $\Phi$, so the reward associated with a trajectory $\xi$ with respect to the $m^{\textrm{th}}$ expert is $R_m(\xi) = \omega_m^\top \Phi(\xi)$, where $\omega_m$ is the unknown vector of weights. Across the expert population, there exists some unknown distribution over the reward parameters, corresponding to the ratio of the data provided by the experts. We represent this distribution with mixing coefficients $\alpha_m$ such that $\sum_{m=1}^M \alpha_m=1$.
We will then learn both the unknown reward functions $\{\omega_m\}_{m=1}^M$ and the mixing coefficients $\{\alpha_m\}_{m=1}^M$, using ranking queries made to the $M$ experts. \begin{revision}This setup generalizes \cite{biyik2020learning}, which studied unimodal rewards.\end{revision}

\noindent\textbf{Ranking Model.}
We define a \emph{ranking query} to be a set of the form $Q\!=\!\qty{\xi_1,\ldots,\xi_K}$ for a fixed query size $K$. The response to a ranking query is a ranking over the items contained therein, of the form $x\!=\!(\xi_{a_1},\ldots, \xi_{a_K})$, where $a_1$ is the index of the expert's top choice, $a_2$ is the second top choice, and so on. While it is not known which expert provided the response to the query, we know the prior that a response comes from expert $m$ with some unknown probability $\alpha_m$, i.e., $\Pr(R=R_m)=\alpha_m$. Going back to our banana placing example, a ranking query of $K$ robot trajectories is generated by the algorithm, and a user---whose identity is unknown to the algorithm---responds to this query.

We then capture how human experts respond to these ranking queries by modeling a ranking distribution through an iterative process using Luce's choice axiom \cite{luce2012individual}. In this process, the experts repeatedly select their top choice $a_m$ with a probability distribution generated with the softmax rule to generate a ranking from the order items were selected:
\begin{align*}
    \Pr\qty(x_1=\xi_{a_1}\mid R=R_m)
     &= \frac
    {e^{R_{m}(\xi_{a_1})}}
    {\sum_{j=1}^K  e^{R_{m}(\xi_{a_j})}}\:.
\end{align*}
In the following iterations, the experts select their top choice among the remaining trajectories:
\begin{align}
    \Pr\qty(x_i=\xi_{a_i}\mid x_1,\ldots, x_{i-1},R=R_m)
     &= \frac
    {e^{R_{m}(\xi_{a_i})}}
    {\sum_{j=i}^K  e^{R_{m}(\xi_{a_j})}}\:.
    \label{eq:conditional}
\end{align}
This is known as the Plackett-Luce ranking model \cite{maystre2015fast,archambeau2012plackett}. Together with the prior over experts $\alpha_m$, the resulting distribution over rankings $x\sim X$ is a mixture of {Plackett-Luce} models with mixing coefficients $\alpha_m$ and weights proportional to $e^{R_m(\xi)}$.

Hence, the ranking distribution first selects the reward function $R_m$ with probability $\alpha_m$, and then selects trajectories from $Q$ sequentially with probability proportional to the exponent of their reward, i.e., $e^{R_m}$, among the remaining trajectories until none is left, generating a ranking of the trajectories.

So given knowledge of the true reward function weights $\omega_m$ and mixing coefficients $\alpha_m$, we have the following joint mass over observations $x$ from a query $Q$:
\begin{equation}
    \label{eq:pmf}
    \Pr(x\mid Q) = \sum_{m=1}^M\alpha_m\prod_{i=1}^K \frac
    {e^{\omega_{m}^\top \Phi(\xi_{a_i})}}
    {\sum_{j=i}^K  e^{\omega_{m}^\top \Phi(\xi_{a_j})}}\:.
\end{equation}
%which will shortly be useful for Bayesian learning in the next section.

%Having presented the problem setup and the ranking model, we are now ready to formally state our goal in this paper.

\begin{comment}
\newcommand{\Cat}[0]{\operatorname{Cat}}
\begin{algorithm}[H]
\caption{Respond to Query}
\begin{algorithmic}[1]
\Require{Query $Q=\qty{\xi_1,\ldots,\xi_K}$}
\State{$R\gets ()$}
\For {$i=1$ to $K$}
\State{$n\sim\Cat(r_1,\ldots)$}
\State{$j\sim \Cat\qty(\qty{e^{\omega_n^\top \Phi(\xi)}:\xi\in Q})$}
\State{$Q\gets Q\setminus\qty{Q_j}$}
\State{$R\gets R\|R_j$}
\EndFor
\State{\Return{$R$}}
\end{algorithmic}
\end{algorithm}
\end{comment}

\noindent\textbf{Objective.} Our goal is to design a series of adaptive queries $Q^{(t)}$ to optimally learn the reward weights $\omega_m$ and corresponding mixing coefficients $\alpha_m$ upon observing the query responses $x^{(t)}$. We constrain all queries to consist of a fixed number of elements $K$.

\section{Active Learning of Multimodal Rewards from Rankings} 

In this section, we first start with presenting our learning framework. We then discuss how we can improve data-efficiency, and propose an active querying approach.

\subsection{Learning from Rankings}
To learn the reward weights $\omega_m$ and mixing coefficients $\alpha_m$, we adopt a Bayesian learning approach. For this, we maintain a posterior over the parameters $\omega_m$ and $\alpha_m$. Denoting the distribution over the parameters $\alpha_i$ and $\omega_i$ as $\Theta$, this posterior can be written as
\begin{revision}
\begin{align}
    \Pr&(\Theta \mid Q^{(1)}, x^{(1)}, Q^{(2)}, x^{(2)}, \dots) \!\propto\! \Pr(\Theta)\Pr(Q^{(1)}, x^{(1)}, Q^{(2)}, x^{(2)}, \dots \mid \Theta)\nonumber\\
    &\hskip -2.2em\!=\!\Pr(\Theta)\prod_{t}\Pr(x^{(t)}, Q^{(t)} \!\mid\! \Theta, Q^{(1)}, x^{(1)} ,\compactdots, Q^{(t-1)}, x^{(t-1)}) \!\propto\! \Pr(\Theta)\prod_{t}\Pr(x^{(t)} \mid \Theta, Q^{(t)}),
    \label{eq:bayesian_learning}
\end{align}
where we use the conditional independence of ranking queries $x^{(t)}$ given $\Theta$ and the conditional independence of the $Q^{(t)}$ on $\Theta$ given $Q^{(1)}, x^{(1)} ,\dots, Q^{(t-1)}, x^{(t-1)}$ in the last equation. \end{revision} To be able to compute this posterior, we assume some prior distribution over the reward weights and the mixing coefficients, which is system-dependent and may come from domain knowledge,  and use Eq.~\eqref{eq:pmf} to calculate the likelihood terms. For example, in our simulations and user studies, we adopted a Gaussian prior $\omega_i\sim\mathcal N(0,I)$ and a uniform prior $\alpha\sim\operatorname{Unif}(\Delta_{M-1})$ where $\Delta_{M-1}$ is the unit $M-1$ simplex. Learning this posterior distribution in Eq.~\eqref{eq:bayesian_learning}, one can compute a maximum likelihood estimate (MLE) or expectation as the predicted reward weights and mixing coefficients.

Equation~\eqref{eq:bayesian_learning} implies the queries made to the experts, $Q^{(t)}$'s, affect how well the posterior will be learned. Assuming a limited budget of queries, which is often the case in many real-world applications, including robotics, one would ideally find an optimal adaptive sequence of queries such that the responses would give the highest amount of information about the reward weights and the mixing coefficients. However, this is NP-hard, even in the unimodal case with pairwise comparisons \cite{ailon2012active}. We therefore resort to greedy optimization techniques to develop our active learning approach.

\subsection{Active Querying via Information Gain}

A query $Q$ is desirable if observing its value $x$ yields high information about the underlying model parameters, $\alpha_m$ and $\omega_m$. Therefore, we propose using an information gain objective to adaptively select the most informative query at each querying step\begin{revision}, generalizing the approach of \cite{biyik2020learning}.\end{revision}% to our more complicated setting\end{revision}.

Assume at a fixed timestep $t$ we have made past query observations $\D=\bigl\{Q^{(t')},x^{(t')}\bigr\}_{t'=1}^{t-1}$. The desired query is then \begin{math}
    Q^*=\arg\max_Q I(X; \Theta \mid Q, \D)
\end{math} where $I(\cdot;\cdot)$ denotes mutual information.  Equivalently, denoting the joint distribution over $x$ and $\theta=\params$ conditioned on $Q$ and $\D$ as $P(X,\Theta\mid Q,\D)$, we see
\begin{align}
    \label{eq:objective}
    Q^*
    &=\argq\min\mathop{\mathbb E}_{P(X,\Theta\mid Q,\D)}\log\frac{\mathop{\mathbb E}_{\theta'\sim\Theta\mid\D}\Pr\qty[X=x\mid Q, \theta']}{\Pr[X=x\mid Q,\theta]}\:.
\end{align} 
The details of this derivation are presented in \Cref{app:derivation}.
%Having formulated the greedy optimization for active querying, we are now ready to present our overall algorithm.

\subsection{Overall Algorithm}
\label{sec:overall}
To efficiently solve the optimization in Eq.~\eqref{eq:objective}, we first note that we should use a Monte Carlo approximation since the expectations are taken over a continuous variable $\Theta$ and a discrete variable $X$ over an intractably large set of $K!$ alternatives. To perform this Monte Carlo integration, we require samples from the posterior $\Pr(X,\Theta\mid Q,\D)$. 

Our key insight is that we can obtain joint samples from both posteriors by first sampling from $\bar\Theta\sim\Pr(\Theta\mid\D)$ and then sampling $x\sim \Pr(X\mid Q,\Theta=\bar\Theta)$ since $\Theta\perp Q \mid\D$ and $X\perp\D\mid Q,\Theta$. We perform the sampling $x\sim \Pr(X\mid Q,\Theta=\bar\Theta)$ efficiently using Eq.~\eqref{eq:pmf}. In general, exact sampling from the posterior $\Pr(\Theta\mid\D)$ is intractable. However, we note Eq.~\eqref{eq:bayesian_learning} can be directly evaluated (using Eq.~\eqref{eq:pmf}) and gives $\Pr(\Theta\mid\D)$ up to a proportionality constant factor.

With this unnormalized posterior of Eq.~\eqref{eq:bayesian_learning}, we use the Metropolis-Hastings algorithm as described in \Cref{app:mh-section} to generate samples from the posterior $\Pr(\Theta\mid\D)$. %Now, these samples are independent of the choice of $Q$. Thus, in our optimization routine for finding $Q$ maximizing Eq.~\eqref{eq:objective}, we can first generate samples $\bar\theta\sim \Pr(\Theta\mid\D)$, and then use these same samples to approximate the objective across different potential queries $Q$.

We see our optimization problem simplifies to finding, for $N$ fixed samples $\bar\theta_i\sim\Pr(\Theta\mid\D)$ and corresponding samples: $x_i\sim\Pr(X\mid Q,\bar\theta_i)$ \begin{align}
    \mathcal{L}(Q; x,\bar\theta) \!=\! \sum_{i=1}^N\!\biggl[\log\Bigl(\sum_{j=1}^N\!\Pr[x_i\mid Q, \bar\theta_j]\!\Bigr)\!-\!\log\Pr[x_i\mid Q,\bar\theta_i]\biggr],\quad
    Q^* = \argq\min \mathcal{L}(Q; x,\bar\theta)\:.
    \label{eq:final}
\end{align}
%Intuitively, viewing each $\bar\theta_i$ as a possible hypothesis over $\params$ (e.g., different driving styles), minimizing each term in the sum in Eq.~\eqref{eq:final} corresponds to selecting $Q$ such that the sampled ranking $x_i$ is maximally unlikely under all hypotheses $\bar\theta_j$ except the hypothesis $\bar\theta_i$ that generated it. In other words, $Q$ is selected so that when its value $x$ is observed, we will have the best possible estimate of which hypothesis in $\bar\theta$ generated it. In the driving example, this means the ranking query will make sure we have a good estimate of which driving style the response came from.

\begin{wrapfigure}{R}{.61\linewidth}
\vspace*{-5ex}
\begin{minipage}{\linewidth}
\begin{algorithm}[H]
\caption{Active Querying via Information Gain}
\def\overi{_{i=1}^N}
\label{alg:info}
\begin{algorithmic}[1]
    \Require{Observations $\mathcal D$}
    \State{$\qty{\bar\theta_i}\overi \! \sim\! \Pr(\Theta\mid\D)$ w/ Eq.~\eqref{eq:bayesian_learning} via Metropolis-Hastings}
    \Procedure{EvalQuery}{$Q$}
    \State{$\forall i, x_i \sim  \Pr(x_i\mid Q,\bar\theta_i)$}
    \State{\Return{$\mathcal L(Q; x,\bar\theta)$}}
    \Comment{Eq.~\eqref{eq:final}}
    \EndProcedure
    \State{$Q\gets \textsc{Minimize}(\textsc{EvalQuery})$}\Comment{Simulated annealing}
    \State{select query $Q$}
\end{algorithmic}
\end{algorithm}
\end{minipage}
\vspace{-20px}
\end{wrapfigure}

We solve this optimization using simulated annealing \cite{bertsimas1993simulated} (see \Cref{app:sa}). Algorithm~\ref{alg:info} goes over the pseudocode of our approach, and we discuss the hyperparameters in our experiments in \cref{app:hyper}.
%We refer to \cref{app:hyper} for the hyperparameters we used in our experiments.

%\paragraph{Simulated Annealing.}
%\label{sec:annealing}
%To optimize $\mathcal L(Q; x, \bar\Theta)$ across $Q$ as in \cref{alg:info}, we use a simulated annealing approach \cite{bertsimas1993simulated} as described in \Cref{app:sa}.

\subsection{Analysis} \label{sec:theory}

We start the analysis by stating the bounds on the required number of trajectories in each ranking query to achieve \emph{generic identifiability}. A Plackett-Luce model over $\Xi$ is generically identifiable if for any sets of parameters $\Theta_1$ and $\Theta_2$ inducing the same distribution over the responses to all queries of size $K$ on $\Xi$, the mixing coefficients of $\Theta_1$ and $\Theta_2$ are the same and the induced rewards ${R_m(\xi)}$ are identical across $\Xi$ up to a constant additive scaling factor.
%(we say $\Theta_1$ and $\Theta_2$ are not \emph{distinct} as parameters). 

\begin{restatable}[\citet{zhao2016learning}]{theorem}{identifiability}
\label{thm:identifiability}
A mixture of $M$ Plackett-Luce models with query size $K$ and $\qty|\Xi|=K$ is {generically identifiable} if $M\leq\left\lfloor K-2\over 2\right\rfloor!$.
\end{restatable}
\vskip -1.5ex
This statement follows directly from \cite{zhao2016learning}, which proves the above bound assuming that each query to the Plackett-Luce mixture is a full ranking over the set of items (i.e. $\qty|\Xi|\!=\!K$). However, the assumption $\qty|\Xi|\!=\!K$ is untenable in the active learning context, as it prevents any active query selection. To apply this result for our active learning algorithms, we relax the condition to $\qty|\Xi|\geq K$.\hskip -1cm
\begin{restatable}{corollary}{largeident}
\label{thm:largeident}
A mixture of $M$ Plackett-Luce models with query size $K$ is \emph{generically identifiable} if $M\leq\left\lfloor K-2\over 2\right\rfloor!$.
\end{restatable}
\vskip -1.5ex
We prove \cref{thm:largeident} in \cref{app:identproof}. In our context, generic identifiability implies if the human response is modelled by a Plackett-Luce mixture, our Algorithm~\ref{alg:info} will be able to recover its true parameters (up to a constant additive factor for the rewards) in the limit of infinite queries.

%\begin{remark}[\citet{zhao2016learning}]
%\label{thm:nonidentifiability}
%A mixture of $M$ Plackett-Luce models over $K$ alternatives is \emph{non-identifiable} if $M\geq\frac{K+1}{2}$, and the bound is tight for $M=2$.
%\end{remark}

%This implies, even in the limit of infinite queries, our algorithm will not be guaranteed to recover the true parameters $\Theta^*$ of the Plackett-Luce mixture model. Notably, if we consider a bimodal Plackett-Luce mixture model with $M=2$, we can guarantee identifiability by using queries of size $K\geq4$. This asserts that pairwise comparison queries, which are commonly used in robotics (e.g., \cite{biyik2020learning,wilde2020active,tucker2020preference}) would not be reliable for multimodal reward learning.

%Next, we discuss the information gain optimization for active querying.
\begin{restatable}{remark}{optimality}
\label{thm:remark}
Greedy selection of queries maximizing information gain in \cref{eq:objective} is not necessarily within a constant factor of optimality.
\end{restatable}
\vskip -1.5ex
\Cref{app:remark} justifies \cref{thm:remark}. In fact, greedy optimization of information gain for adaptive active learning can be significantly worse than a constant factor of optimality in pathological settings \cite{golovin2010near}. Despite its lack of theoretical guarantees, information gain is a commonly used effective approach in adaptive active learning \cite{biyik2019asking,houlsby2011bayesian,zheng2005efficient}. Although other approaches like volume removal satisfy adaptive submodularity \cite{sadigh2017active}, they fail in settings with noisy observations by selecting high-noise low-information queries, and in practice achieve far worse performance than information gain.

\vspace{-2px}
\section{Experiments}
\vspace{-2px}
\label{sec:experiments}
\begin{wrapfigure}{R}{.5\textwidth}
    \centering
    \vspace{-25px}
    \includegraphics[width=\linewidth]{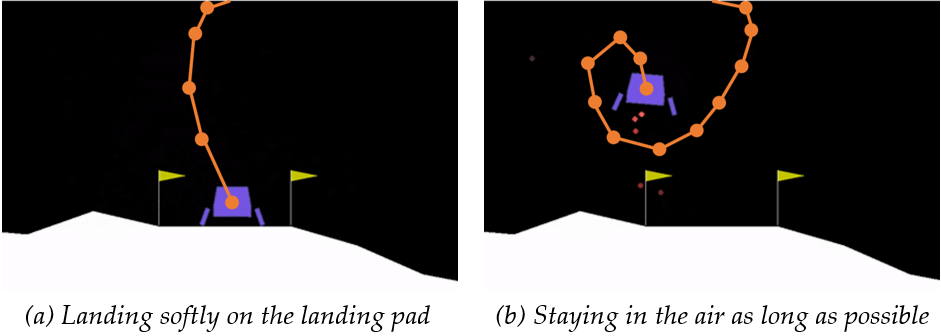}
    \vspace*{-15px}
    \caption{The LunarLander environment is visualized with the two tasks. Sample trajectories associated with these tasks are shown.}
    \vspace*{-10px}
    \label{fig:lunarlander_visual}
\end{wrapfigure}
Having presented our learning and active querying algorithms, we now evaluate their performance in comparison with other alternatives. We start with describing the two tasks we experimented with:\\
\noindent\textbf{LunarLander.} We used 1000 trajectories in OpenAI Gym's LunarLander environment~\cite{brockman2016openai} shown in Fig.~\ref{fig:lunarlander_visual} (see \Cref{app:lunar-traj} for details on how they were generated).\\
\noindent\textbf{Fetch Robot.} We generated $351$ distinct trajectories of the Fetch robot~\cite{wise2016fetch} putting the banana on the shelves as shown in Fig.~\ref{fig:robot_experiment_visual} (see \Cref{app:fetch-traj} for details).

\vspace{-2px}
\subsection{Methods}\label{sec:benchmark_methods}
\vspace{-2px}
% We now introduce the methods we apply in our experiments and user studies, where we compare the benchmarks with the active querying via information gain (IG) method we proposed in Algorithm~\ref{alg:info}.

We compare our active querying via information gain (IG) discussed in Algorithm~\ref{alg:info} with two baselines:
A simple benchmark for active learning is \emph{random} query selection without replacement. We also benchmark against \emph{volume removal} (VR), a common objective for active learning of robot reward functions \cite{sadigh2017active}. See \Cref{app:baselines} for the details of these two baselines.

\subsection{Metrics}

We want to evaluate both the active querying and the learning performance. The former requires metrics that assess the quality of the algorithm's selected queries $\D\!=\!\qty{(Q^{(t)},x^{(t)})}_t$ in terms of the information they provide on the model parameters $\Theta$. We use two such metrics: mean squared error (MSE) and log-likelihood. Since both active and non-active methods are expected to reach the same performance with a large number of queries, we look at the area under the curve (AUC) of these two metrics over number of queries. To evaluate the learning performance, we quantify the success of a robot, which learned a multimodal reward, via the \emph{learned policy rewards} on the actual task.

\paragraph{MSE.} Suppose we know the human is truly modeled by $\Theta^*$ adhering to the assumed model class of \cref{sec:formulation}. Given a set of observations $\D$, we can compute an MLE estimate $\widehat\Theta$ of the model parameters using \cref{eq:pmf}. The MSE is then the squared error between $\widehat \Theta$ and $\Theta^*
$ (see \Cref{app:mse}).

While this metric cannot be evaluated with real humans, we can use this metric with synthetic human models (model with known parameters $\Theta^*$) in simulation. A lower MSE score means the selected queries $\D$ allow us to better learn a multimodal Plackett-Luce model close to the true model $\Theta^*$.

\paragraph{Log-Likelihood.}
The log-likelihood metric measures the log-likelihood of the response to a random query given the past observations $\D$. If the past observations $\D$ are informative, the true response to a random query $Q$ will in expectation be more likely, meaning the log-likelihood metric will be greater. See \Cref{app:ll} for details on how we compute this metric.

\paragraph{Learned Policy Reward.}
We take the MLE estimate of each reward weights vector and train a DQN policy using them \cite{mnih2013playing}.\footnote{As we are using a real Fetch robot for our experiments and it would be infeasible and unsafe to train DQN on Fetch, so this metric is limited to our simulations, i.e., LunarLander in our experiments.} We then run these learned policies on the actual environment with the corresponding true reward functions (see \Cref{app:learned_policy_rewards}) to obtain the learned policy rewards.

\subsection{Results}

\paragraph{Multimodal Learning is Necessary.} We first compare unimodal and multimodal models to show the insufficiency of unimodal rewards when the data come from a mixture. To leave out any possible bias due to active querying, we make this comparison using random querying.

\begin{wrapfigure}{R}{.5\textwidth}
    \centering
    \vspace{-20px}
    \includegraphics[width=\linewidth]{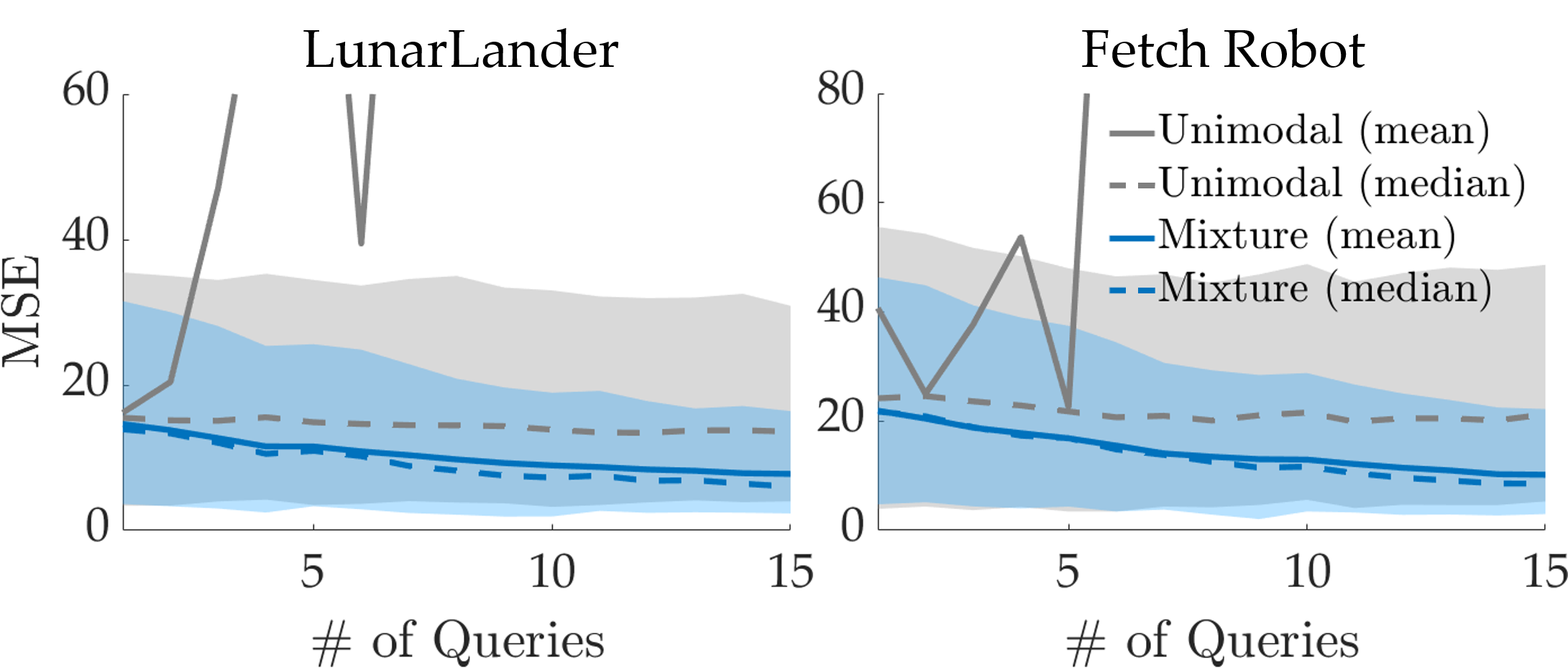}
    \vspace{-15px}
    \caption{Unimodal and bimodal reward learning models are compared under MSE. Both mean and median values (over $100$ runs) are shown. %, because the unimodal model has many outliers that cause the mean to diverge. 
    Shaded regions show the first and the third quartiles.
    %Unimodal model is not able to learn the reward, and so the MSE values do not decrease. On the other hand, our multimodal reward model enables learning from mixture data.
    }
    \label{fig:unimodal_vs_bimodal}
    \vspace{-20px}
\end{wrapfigure}

We let the true reward function have $M\!=\!2$ modes and set a query size of $K\!=\!6$ items for identifiability as Section~\ref{sec:theory} suggests, and for acquiring high information from each query. We simulate $100$ pairs of experts whose reward weights $\omega_m$ and the mixing coefficients $\alpha_m$ are sampled from the prior $\Pr(\Theta)$. Having these simulated experts respond to $15$ queries, we report the MSE in \cref{fig:unimodal_vs_bimodal}.

The unimodal reward model causes an unstably increasing MSE. This is mostly due to the outliers where the reward weights $\omega_1$ and $\omega_2$ are far away from each other and the unimodal reward fails to learn any of them. We therefore also plot the median values and quartiles in Fig.~\ref{fig:unimodal_vs_bimodal}. While the bimodal reward model learned using our proposed approach decreases the MSE over time, the unimodal model has a roughly constant MSE, which suggests it is unable to learn when the data come from a mixture.

\begin{revision}
We present an additional unimodal learning baseline evaluated on the user study data in \Cref{app:new_baseline}.
\end{revision}

\begin{wrapfigure}{R}{.5\textwidth}
    \centering
    \vspace{-18px}
    \includegraphics[width=\linewidth]{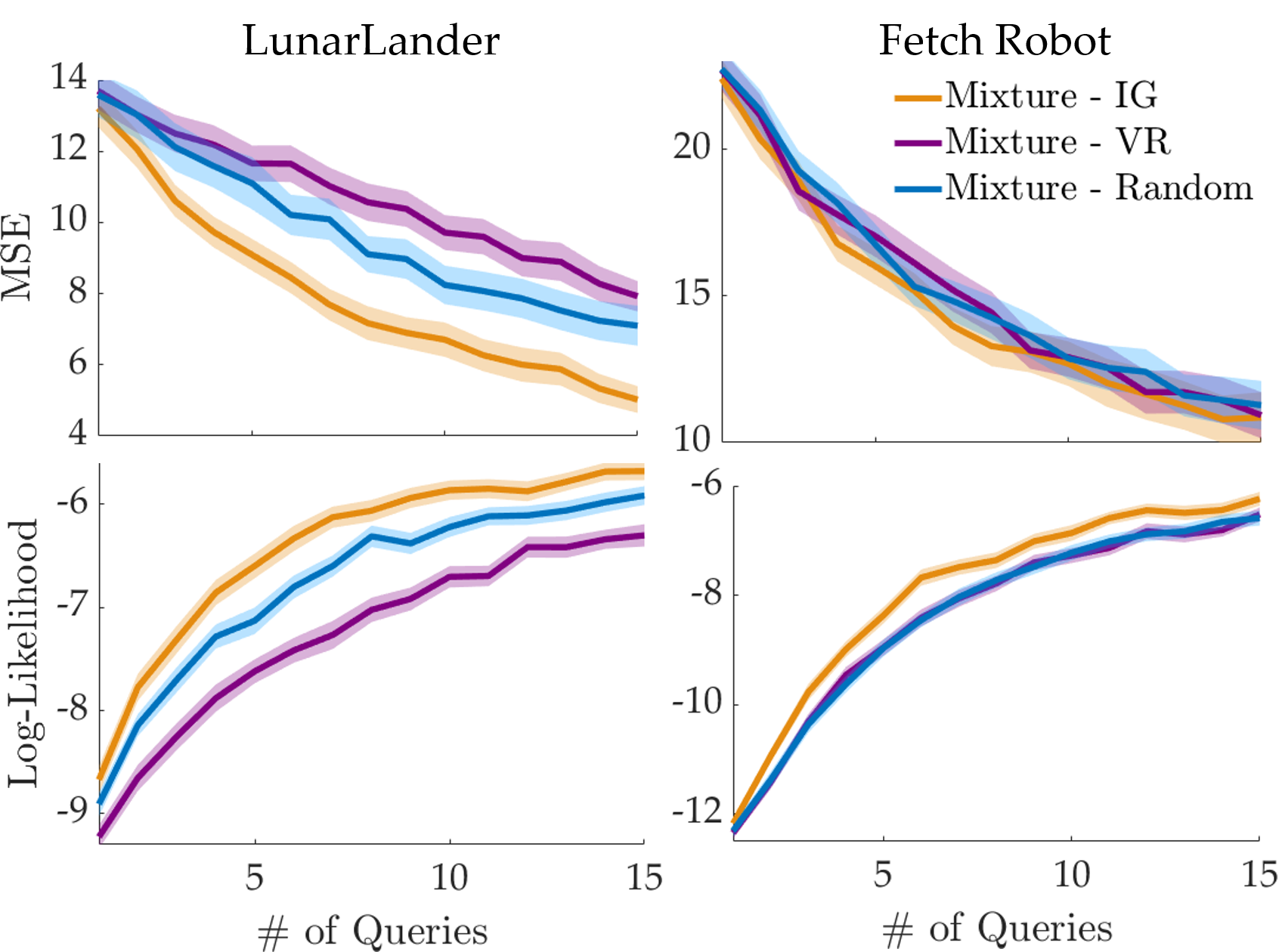}
    \vspace*{-15px}
    \caption{Different querying methods are compared with the (top) MSE and (bottom) log-likelihood metrics (mean$\pm$se over $75$ runs).}
    \vspace*{-20px}
    \label{fig:bimodal}
\end{wrapfigure}

\paragraph{Active Querying with Information Gain is Data-Efficient.} We next compare our information-gain-based active querying approach with the other baselines. For this, we use the same experiment setup as above with $M=2$ reward function modes and ranking queries of size $K=6$,  and simulate $75$ pairs of human experts. We present the results in terms of MSE in Fig.~\ref{fig:bimodal}. In LunarLander, the information gain objective significantly outperforms both random querying and volume removal in terms of the AUC MSE ($p<0.005$, paired-sample $t$-test). Notably, volume removal performs even worse than the random querying method, which might be due to the known issues of volume removal optimization as briefly discussed in \cref{app:volume}.
On the other hand, the difference is not statistically significant in the Fetch Robot experiment, which might be due to the small trajectory dataset, or because almost all trajectories in the dataset minimize or maximize some of the trajectory features, accelerating and simplifying learning under the linear reward assumption. See Appendix~\ref{app:fetch-traj} for details about the trajectory features and how we generated the trajectory dataset.

We further analyze the querying methods in this multimodal setting under the log-likelihood metric in Fig.~\ref{fig:bimodal}. Information gain significantly outperforms random querying and volume removal in both experiments with respect to the AUC log-likelihood ($p<0.005$).
With respect to the final log-likelihood, information gain reduces the amount of required data in LunarLander by about $35\%$ compared to random querying and about $60\%$ to volume removal. Similarly in the Fetch Robot, the improvement is approximately $25\%$ over both baselines.

\cref{app:syn} presents two additional experiments: one which clearly shows the effectiveness of our approach for learning a mixture of more than two reward functions (specifically, $M\!=\!5$), and one which studies the robustness against misspecified $M$.

\paragraph{Information Gain Leads to Better Learning.} Having seen the superior predictive performance of the reward learned via information gain optimization, we next assess its performance in the actual environment. As random querying outperforms volume removal in terms of log likelihood and MSE as in Fig.~\ref{fig:bimodal}, we compare the information gain with random querying.

\begin{wrapfigure}{R}{.5\textwidth}
    \centering
    \vspace*{-30px}
    \includegraphics[width=\linewidth]{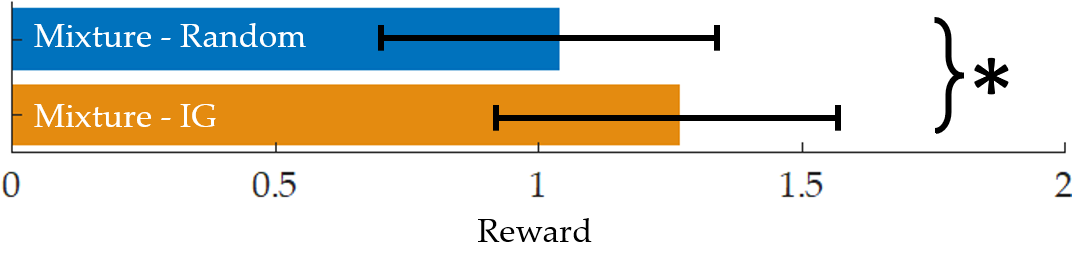}
    \vspace{-18px}
    \caption{Information gain and random querying methods are compared with the learned policy rewards (mean$\pm$se over $75$ runs which correspond to $150$ randomly generated reward weights) in LunarLander.}
    \vspace*{-8px}
    \label{fig:bimodal_dqn}
\end{wrapfigure}

For this, we run the multimodal reward learning with $75$ pairs of randomly generated reward weights ($M=2$ and $K=6$). For each of the $150$ individual reward functions, we compute the learned policy rewards. Fig.~\ref{fig:bimodal_dqn} shows the results. While the standard errors in the plots seem high, this is mostly because optimal trajectories for different reward weights differ substantially in terms of rewards, which causes an irreducible variance. However, since the underlying true rewards are the same between the information gain and random querying methods, we ran the paired sample $t$-test between the results and observed statistical significance ($p<0.05$). This means although the learned policy rewards between different runs differ substantially, the reward function learned via the information gain method leads to better task performance compared to random querying.

\vspace{-2px}
\section{User Studies}
\vspace{-2px}

We now empirically analyze the performance of our algorithm with two online user studies.\footnote{We have IRB approval from a research compliance office under the protocol number IRB-52441.} We again used the LunarLander and Fetch Robot environments. We provide a summary and a video of the user studies and their results at \url{https://sites.google.com/view/multimodal-reward-learning/}.

\paragraph{Experimental Setup.} For LunarLander, subjects were presented with either of the following instructions at every ranking query: ``Land softly on the landing pad" or ``Stay in the air as long as possible". We randomized these instructions such that users get one of them with $0.6$ and the other with $0.4$ probability%
%(instruction-probability matching was done uniformly at random)
. We kept the presented instructions hidden from the learning algorithms so that they need to learn a multimodal reward without knowing which mode each ranking belongs to.

For the Fetch robot environment, we recorded the $351$ trajectories on the real robot as short video clips so that the experiment can be conducted online under the pandemic regulations. Human subjects participated in the experiment as groups of two to test learning from multiple users. Each participant was instructed that the robot needs to put the banana in one of the shelves and different shelves have different conditions (the same as in our running example, see Fig.~\ref{fig:robot_experiment_visual}, \cref{app:desc}).

After emphasizing there is no one correct choice and it only depends on their preferences, we asked each participant to indicate their preferences between the shelves on an online form. Afterwards, each group of two subjects responded to $30$ ranking queries in total where each query consisted of $6$ trajectories. 
%To prevent possible shifts in the subjects' preferences due to their partner in the experiment, we did not let them see the initial preferences or responses to ranking queries of each other. 
We selected who responds to each query randomly, with probabilities $0.6$ and $0.4$. 
%We kept the identities of the subjects responding to queries hidden from the learning algorithms so that they would need to learn a multimodal reward without knowing which mode each ranking belongs to.

\cref{app:ui} presents details on the user interface used in our experiments.

\paragraph{Independent Variables.} We varied the querying algorithm: active with information gain and random querying. We excluded the volume removal method to reduce the experiment completion time for the subjects, as it already performed worse than random querying in our simulations.

\paragraph{Procedure.} We conducted the experiments as a within-subjects study. We recruited $24$ participants (ages 19 -- 56; $9$ female, $15$ male) for LunarLander and $26$ participants (ages 19 -- 56; $11$ female, $15$ male) for the Fetch robot. Each subject in the LunarLander, and each group of two subjects in the Fetch robot experiment responded to $40$ ranking queries; $15$ with each algorithm and $10$ random queries for evaluation at the end. The order of the first $30$ queries was randomized to prevent bias. %The participants knew the number of queries in advance. 
%Based on the forms where participants indicated their preferences between the shelves in the Fetch robot experiments, $11$ of the $13$ groups consisted of participants with different preferences.
%While the participants had the same preferred shelf order in the other $2$ groups, their preferences about how fast the robot should move, where it should grasp the banana, and where it should place in a shelf may have been different.

\paragraph{Dependent Measures.} Learning the multimodal reward functions via the $15$ rankings collected by each algorithm, we measured the log-likelihood of the final $10$ rankings collected for evaluation.

\paragraph{Hypotheses.} With LunarLander and Fetch robot, we test the following hypotheses respectively:\\
    \textbf{H1.} \textit{Querying the participants, who are trying to teach two different tasks, actively with information gain will lead to faster learning than random querying.}\\
    \textbf{H2.} \textit{While learning from two people with different preferences, active querying with information gain will lead to faster learning than random querying.}

\begin{wrapfigure}{R}{.5\textwidth}
    \centering
    \vspace{-20px}
    \includegraphics[width=\linewidth]{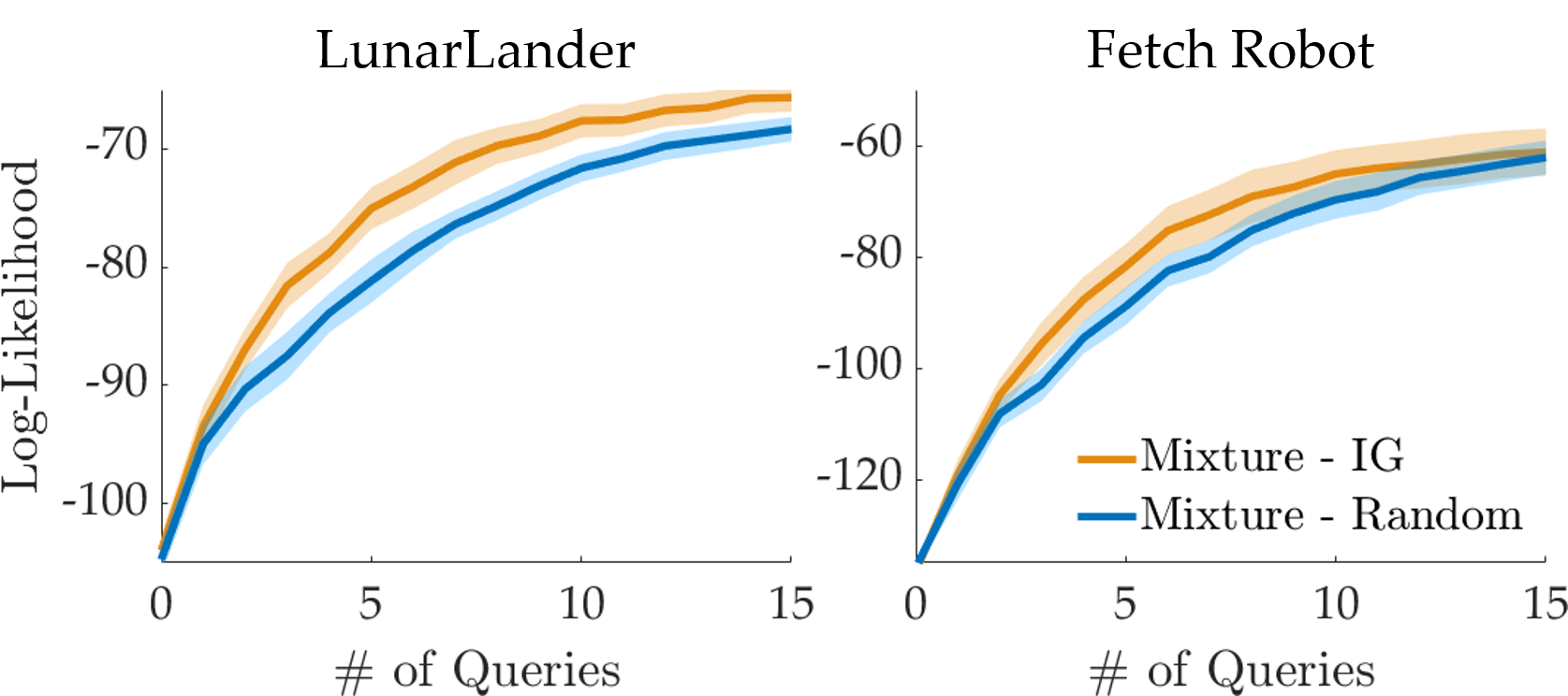}
    \vspace*{-15px}
    \caption{User study results (mean$\pm$se over $24$ users for LunarLander and $13$ groups for Fetch Robot).}
    \vspace{-15px}
    \label{fig:user_studies_loglikelihood}
\end{wrapfigure}

\paragraph{Results.} Figure~\ref{fig:user_studies_loglikelihood} visualizes how log-likelihood of the evaluation queries changes over the course of learning by both algorithms. Active querying with information gain leads to significantly faster learning compared to random querying in LunarLander. 
%By the end of $15$ queries, the active method achieves a log-likelihood of $-65.7\pm1.2$ whereas random querying is at $-68.3\pm1.0$. 
%This means the multimodal reward learned with the active method assigned a $14.4$ times higher probability on the evaluation rankings than the random querying method (from the log-likelihood metric). 
Indeed, the difference in AUC log-likelihood is statistically significant ($p<0.05$). Furthermore, the active querying method enabled reaching the final performance of random querying after only $9$ or $10$ queries, for around a $35\%$ reduction in the amount of data needed, supporting \textbf{H1}.

As the robot experiments have a an easier task with a small number of variables between the trajectories, both querying methods converge to similar performances by the end of $15$ queries. However, active querying accelerates learning in the early stages%
%By the end of $15$ queries, the active and random querying methods achieve log-likelihoods of $-61.2\pm4.4$ and $-62.2\pm3.2$, respectively, corresponding to a $2.7$x increase in the likelihood of evaluation queries
---the difference in AUC log-likelihood is again statistically significant ($p<0.05$). Looking at the final performance with random querying, improvement in data efficiency is about $10\%$, supporting \textbf{H2}.

\vspace{-2px}
\section{Conclusion}
\vspace{-2px}

\paragraph{Summary.} This work presents a novel approach for learning multimodal reward functions. We formulated the problem as a mixture learning problem solved using ranking queries that are answered by experts. We further developed an active querying method that maximizes information gain to improve the quality of ranking queries made to the experts. The results suggest our model learns multimodal reward functions, with data efficiency improved by our new active querying method.

% \smallskip
\paragraph{Limitations and Future Work.} Our model for learning multimodal rewards requires knowing the number of different modes (experts or tasks) $M$ in advance. This might be difficult in some settings. For example, when several experts belonging to different clusters, e.g., timid and aggressive drivers, provide data, it might be difficult to know the number of clusters in advance. However, a simple approach that fits the multimodal reward under various $M$ could reveal the true number of underlying modes. Another challenge is that learning a mixture reward model may contribute to the reward ambiguity problem in inverse reinforcement learning: each individual reward may have its own ambiguity. Future work should investigate the practical implications of this. In addition, theoretical results assert, to guarantee the reliable learning of a multimodal reward with $M$ modes, ranking queries should consist of $K$ queries such that $M \leq \lfloor\frac{K-2}{2}\rfloor!$. While this is manageable by multiple pairwise comparisons for each query or an iterative process where the expert selects the top item, it might consume too much time for large $M$. Thus, an interesting future direction is to investigate how to incorporate multiple forms of expert feedback, e.g., demonstrations in addition to rankings, to pretrain and reduce the required interaction time with humans.

% The acknowledgments are automatically included only in the final and preprint versions of the paper.
\acknowledgments{The authors would like to acknowledge funding by NSF grants \#1849952 and \#1941722, FLI grant RFP2-000, and DARPA.}

%===============================================================================

\setlength{\bibsep}{4.40pt plus 0.3ex}
{\small\bibliography{references}}  % .bib

\newpage
\appendix

In the appendix, we provide additional details on the derivations of our methods and analysis, in addition to methodological details omitted from the main paper. In \Cref{app:derivation}, we directly derive the formulas necessary for our querying via information gain optimization approach. \Cref{app:mh-section,app:sa,app:hyper} present details on our main approach in \Cref{alg:info}. \Cref{app:proofs} provides the arguments needed to justify the claims of \Cref{sec:theory}. Finally, \Cref{app:baselines,app:traj,app:metrics,app:expset} present details on our experimental setups, while \Cref{app:syn} presents a compelling additional synthetic data experiment demonstrating learning a reward function mixture with five modes.

\section{Information Gain Derivation}
\label{app:derivation}
We present the derivation of the formula for computing the maximum information gain query $Q^*$.
Assume at a fixed timestep $t$ we have made past query observations $\D=\{Q^{(t')},x^{(t')}\}_{t'=1}^{t-1}$. The desired query is then 
\begin{align}
\label{eq:initial_objective}
    Q^*=\argq\max I(X; \Theta \mid Q, \D),
\end{align} 

where $I(\cdot;\cdot)$ denotes mutual information. Equivalently, denoting conditional entropy with $\mathcal H[\cdot\mid\cdot]$, we note
\begin{align*}
    I(X; \Theta \mid Q, \D) \!&= \H\qty[\params\!\mid \D] - \mathop{\mathbb E}_{P(X\mid Q, \D)}\!\bigg[ \H\qty[\params \!\mid\! Q,X\!=x,\D]\bigg]\:,
\end{align*}

which allows us to write the optimization in \Cref{eq:initial_objective} equivalently as
\begin{align*}
    Q^* &= \argq\min \mathop{\mathbb E}_{P(X\mid Q, \D)}\!\bigg[\H\qty[\params\mid Q,X=x,\D]\bigg]\:.
\end{align*}

We further simplify this minimization objective by denoting the joint distribution over $x$ and $\theta=\params$ conditioned on $Q$ and $\D$ as $P(X,\Theta\mid Q,\D)$ and expanding the entropy term:

\begin{align*}
    Q^*&=\argq\min\mathop{\mathbb E}_{P(X,\Theta\mid Q, \D)}\log\frac{\Pr\qty[X=x\mid Q, \D]}{\Pr[X=x\mid Q,\theta]}\nonumber\\
    &=\argq\min\mathop{\mathbb E}_{P(X,\Theta\mid Q,\D)}\log\frac{\mathop{\mathbb E}_{\theta'\sim\Theta\mid\D}\Pr\qty[X=x\mid Q, \theta']}{\Pr[X=x\mid Q,\theta]}\:. \tag{see~\ref{eq:objective}}
\end{align*}

\section{Metropolis-Hastings} \label{app:mh-section}
\subsection{Approach}
\label{app:mh-methods}

\newbox\metrofigure
\begin{wrapfigure}{R}{.49\textwidth}
\vspace*{-2.5ex}
\global\setbox\metrofigure=\vbox{
    \centering
    \includegraphics[width=\linewidth]{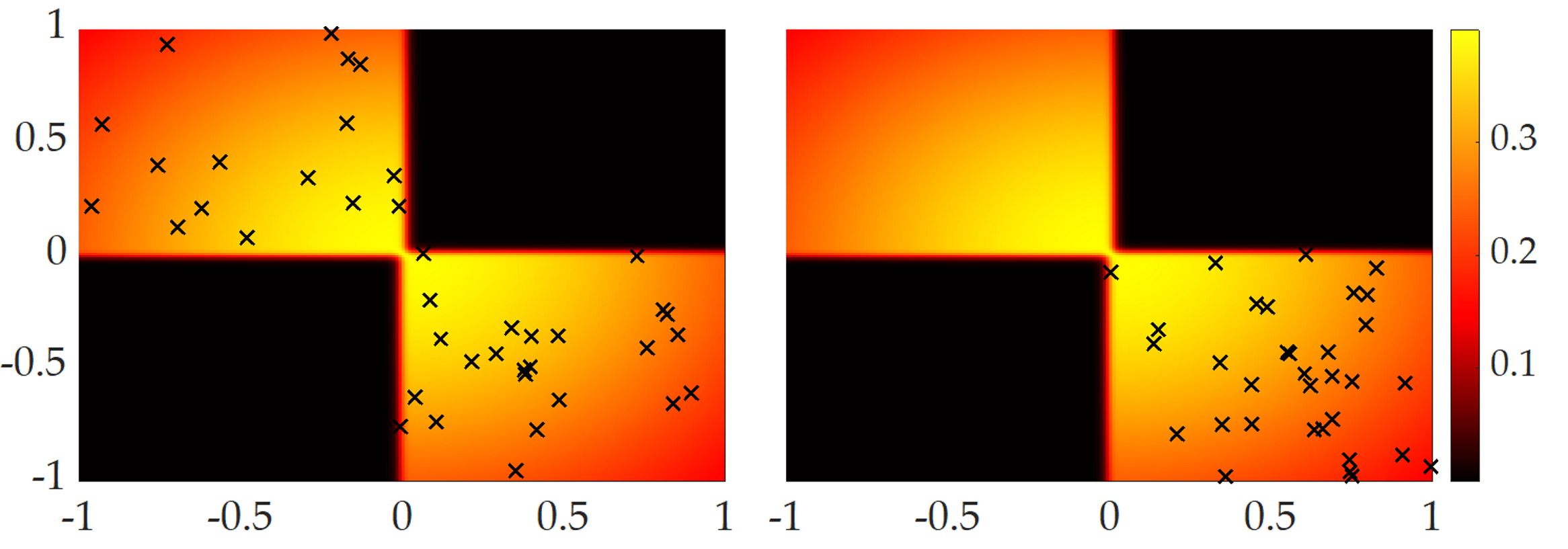}
    % \vspace*{-15px}
    \caption{Multi-chain Metropolis-Hastings sampling \textbf{(left)} gives more representative samples from the distribution compared to the single-chain variant \textbf{(right)}.}
    \label{fig:MH_sampling}
    }
\copy\metrofigure
\vspace{-2.5ex}
\end{wrapfigure}

To sample from $\Pr(\Theta\mid\D)$ using \cref{eq:bayesian_learning}, we use the Metropolis-Hastings algorithm \cite{chib1995understanding}, running $N$ chains simultaneously for $H_{MH}$ iterations.
To avoid autocorrelation between samples, unlike in conventional Metropolis-Hastings we only use the last state in each chain as a sample.
In contrast, for conventional Metropolis-Hastings, multiple samples would be drawn from a single chain at set intervals after a short \emph{burn-in} period. As we see in \cref{fig:MH_sampling}, for our multimodal Plackett-Luce posteriors, performing multi-chain Metropolis-Hastings yields posterior samples that are far more evenly distributed across different posterior modes. Thus, to achieve well-distributed posterior samples, we set our effective burn-in period to be $H_{MH}-1$, taking only the last sample from each chain.

For two states in the chain $\Theta=\params$ and $\Theta'=\params[|']$, our proposal distribution is then $$g(\Theta'\mid\Theta) = \prod_{m=1}^M \varphi(\omega_m - \omega_m'),$$ where $\varphi$ is the pdf of the diagonal Gaussian $\mathcal N(0, \sigma_{MH} I)$.

\subsection{Multimodal Metropolis-Hastings Demonstration}
\label{app:metropolis}

The posterior distribution in the figure is that of a 2-mode Plackett-Luce mixture with fixed uniform mixing coefficients and 1-D weights conditioned on the observations $50\succ -50$ and $-50\succ 50$. The single-chain algorithm ran for 2000 steps with a burn-in period of 200 steps after which every $18$th sample was selected, while the multi-chain algorithm used 100 chains for 20 iterations each, taking only the last sample from each chain.

\section{Simulated Annealing}\label{app:sa}
For our simulated annealing, we run $N_{SA}$ chains in parallel for $H_{SA}$ iterations each, returning the best query $Q$ found across each run. We define the transition proposal distribution $g(Q'\mid Q)$ to be a positive constant if $Q'$ and $Q$ differ by one trajectory and $0$ otherwise. We run with a starting temperature of $T^0_{SA}$, cooling by a factor of $\gamma_{SA}$ with each subsequent iteration past the first.

\section{Hyperparameters}
\label{app:hyper}

We use the hyperparameters in \cref{table:hyper} for the simulated annealing and Metropolis-Hastings algorithms, whose details are provided in \cref{app:mh-section} and \cref{app:sa}, respectively.
\begin{table}[H]
\vspace{-20px}
\caption{Hyperparameters}
\smallskip
\begin{displaymath}
\begin{array}{cc}
    \toprule
    \makebox[5em]{Constant} & \makebox[5em]{Value} \\ \midrule
    \rule{0pt}{2ex} 
     N & 100 \\
     H_{MH} &  200 \\
     \sigma_{MH} & 0.15 \\
     N_{SA} & 10 \\
     H_{SA} & 30 \\
     T^0_{SA} & 10 \\
     \gamma_{SA} & 0.9\\\bottomrule
\end{array}
\end{displaymath}
\label{table:hyper}
\vspace{-20px}
\end{table}

\section{Proofs and Analysis}
\label{app:proofs}
\subsection{Proof of \Cref{thm:largeident}}
\label{app:identproof}
\largeident*
\vskip -1.5ex
Suppose we have such a mixture of $M$ Plackett-Luce models that is not identifiable. Then, there must exist two distinct sets of parameters $\Theta_1$ and $\Theta_2$ such that for every query $Q$, the induced ranking distributions $X_1$ and $X_2$ respectively are identical. But since $\Theta_1$ and $\Theta_2$ are distinct, there is either (1) two mixing coefficients in $\Theta_1$ and $\Theta_2$ that disagree or (2) two items $\xi_1$ and $\xi_2$ that have a different difference in rewards across $\Theta_1$ and $\Theta_2$ under one of the reward functions. Let $\bar Q$ with corresponding ranking distribution $\bar X$ be an arbitrary query in case (1) and an arbitrary query containing $\xi_1$ and $\xi_2$ in case (2). Note that $\bar X$ is the marginal distribution of the overall Plackett-Luce distribution, which by construction is a mixture of $M$ Plackett-Luce models with parameters $\Theta_1$ and $\Theta_2$, restricted to the items in $\bar Q$. But now there are two distinct sets of parameters representing the distribution over the full ranking of $Q$ since we know $\Theta_1$ and $\Theta_2$ differ on the restricted set of items $\Xi'=Q$ (either because they have differing mixing coefficients or because their induced rewards on $\Xi'$ are not a within a constant additive factor of each other since $\xi_1$ and $\xi_2$ are in $\Xi'$). But we know $\qty|\Xi'|=\qty|Q|=K$, so this finding contradicts the fact that $\bar X$ must be identifiable by \cref{thm:identifiability}. We conclude every mixture of $M$ Plackett-Luce models is identifiable subject to the query size bounds in the statement of this corollary.\qed

\subsection{Justification for \Cref{thm:remark}}
\label{app:remark}
\optimality*
\vskip -1.5ex
Here, we define the optimal adaptive set of queries $\D$ to be the one which, in expectation, minimizes the uncertainty over model parameters $\mathcal H\qty[\Theta \mid \D]$. It is a well-known result that for \emph{adaptive submodular} functions, greedy optimization yields results that are within a constant factor $\qty(1-\frac1e)$ of optimality \cite{golovin2011adaptive}. While our mutual information objective in Eq.~\eqref{eq:objective} is adaptive submodular in the non-adaptive setting (where all queries $Q$ are selected before observing their results), in our adaptive setting these guarantees no longer hold (conditional entropy is only submodular with respect to conditioned variables if those variables are unobserved).

\section{Baselines} \label{app:baselines}
\subsection{Random}
\label{app:random}
We benchmark against a random agent, wherein at each step the query selected by the agent is a collection of $K$ random items without replacement. We also use the random querying method for comparing the multimodal reward learning with the existing approaches that assume a unimodal reward (e.g. \cite{biyik2020learning}), as it does not introduce any bias in the query selection.
\subsection{Volume Removal}
\label{app:volume}
Volume removal seeks to maximize the difference between the prior distribution over model parameters and the \textit{unnormalized} posterior. Volume removal notably fails to be optimal in domains where there are similar trajectories \cite{biyik2019asking}. In these settings, querying sets of trajectories with similar features removes a large amount of volume from the unnormalized posterior (since the robot is highly uncertain about their relative quality), yet yields little information about the model parameters (since the human also has high uncertainty). Information gain approaches such as our method are better able to generate trajectories to query for which the robot has high uncertainty while the human has enough certainty to yield useful information for the robot.

\section{Trajectory Generation}\label{app:traj}
\subsection{LunarLander Trajectories}\label{app:lunar-traj}
\begin{figure}[H]
    \centering
    \begin{minipage}{.45\linewidth}
    \centering
    \includegraphics[width=\linewidth]{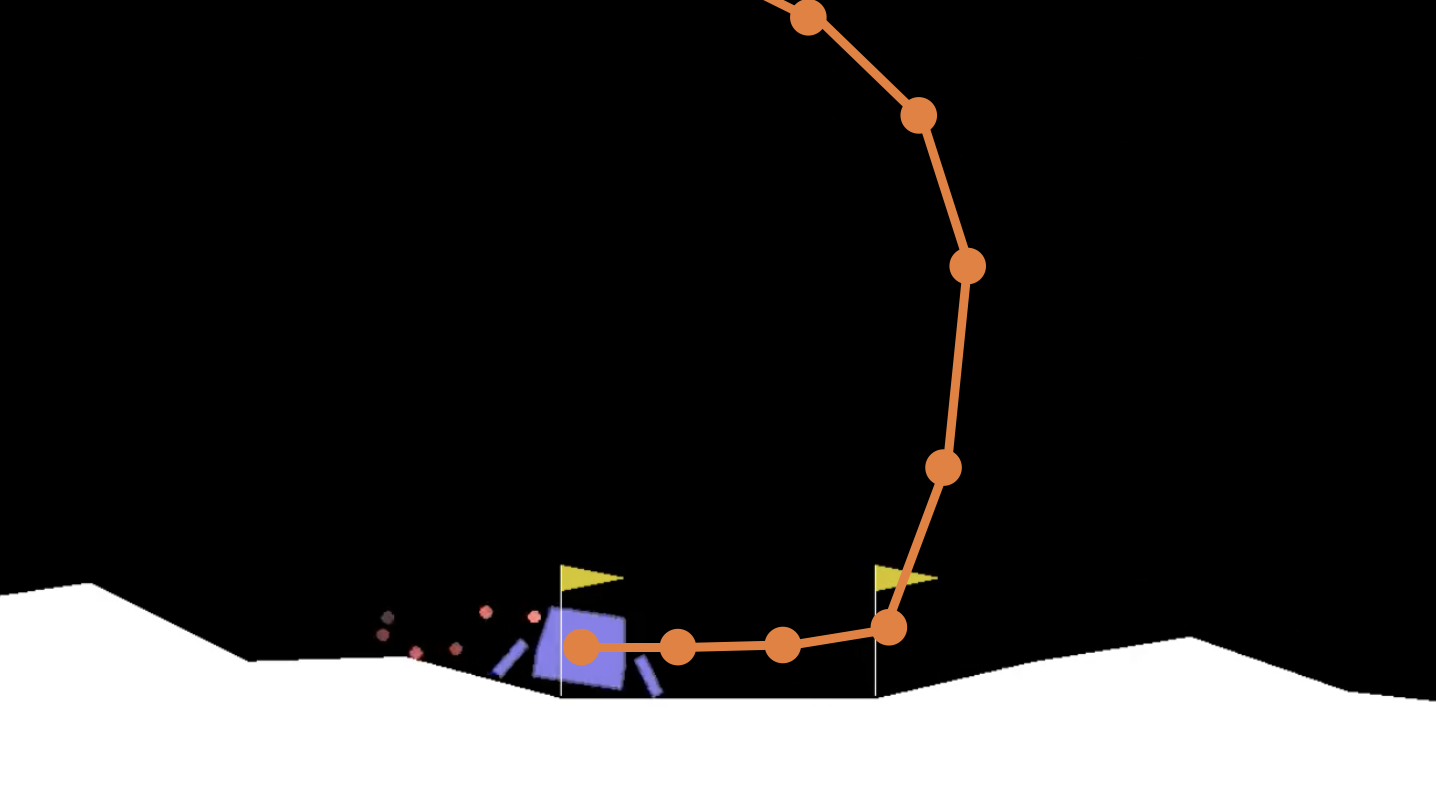}
    \end{minipage}\hskip 2em\begin{minipage}{.45\linewidth}
    \centering
    \begin{tabular}{cr}
    \toprule
         Feature & \multicolumn{1}{c}{Value} \\
     \midrule
        Mean angle & $2.27683634$   \\
        Total angle & $-0.20375356$ \\
        Distance to Goal & $5.41860642$ \\
        Total rotation & $0.25948072$ \\
        Path length & $3.71660086$ \\
        Final vertical velocity & $-0.57097337$ \\
        Crash time & $1.11112885$ \\
        Score & $-0.15500268$ \\
     \bottomrule
    \end{tabular}
    \end{minipage}
    \caption{Sample LunarLander trajectory (left) with extracted features (right).}
    \label{fig:lunartraj}
\end{figure}

We designed $8$ trajectory features based on: absolute heading angle accumulated over trajectory, final distance to the landing pad, total amount of rotation, path length, task completion (or failure) time, final vertical velocity, whether the lander landed on the landing pad without its body touching the ground, and original environment reward from OpenAI Gym. Using these features, we randomly generated $10$ distinct reward functions based on the linear reward model and trained a DQN policy \cite{mnih2013playing} for each reward. Finally, we generated $100$ trajectories by following each of these $10$ policies in the environment to obtain $1000$ trajectories in total. We used these trajectories as our dataset for the ranking queries. \Cref{fig:lunartraj} presents an example trajectory with extracted scaled and centered features.

\subsection{Fetch Trajectories}\label{app:fetch-traj}
To design our $351$ trajectories, we varied the target shelf (3 variations), the movement speed (3), the grasp point on the banana (3) and where in the shelf it is placed (13). We then designed $12$ trajectory features based on these varied parameters and appended another binary feature which indicates whether any object dropped from the shelves on that trajectory.

\def\succ{y_{\text{success}}}
\begingroup
\def\speed{y_{\text{speed}}}
\def\grasp{y_{\text{grasp}}}
\def\height{y_{\text{height}}}
\def\width{y_{\text{width}}}
Specifically, for $\tau$ a trajectory, let $$x_i=
\left\{\smqty{1 & ~~\makebox[8em][l]{$i$ is the target shelf}\\
0&~~\makebox[8em][l]{otherwise}}\right.,$$

$\grasp,\height,\width,\speed$ specify the grasp position and speed, and $\succ$ specify whether the robot did not drop any objects from the shelves. Our featurization is then
\begin{align*}\Phi(\tau)&=\bigl( x_1,x_2,x_3,\speed,\speed(1-\speed),\grasp,\\&\qquad\grasp(1-\grasp),\height,\height(1-\height),\\&\qquad\width, \width(1-\width),1-(\grasp-\width)^2,\succ \bigr) .\end{align*} \Cref{fig:fetchtraj} presents a sample Fetch trajectory with its featurization.
\endgroup

\xdef\x{1.0000, 0.0000, 0.0000, 0.5000, 0.2500, 1.0000, 0.0000, 0.7500, 0.1875,
        0.2500, 0.1875, 0.4375, 1.0000,}
\def\first#1,#2\end{#1}
\def\rest#1,#2\end{#2}
\def\popx{
$\expandafter\first\x\end$
\xdef\x{\expandafter\rest\x\end}
}

\def\htt{y_{\rm height}}
\def\wtt{y_{\rm width}}
\begin{figure}[H]
    \centering
    \begin{minipage}{.45\linewidth}
    \centering
    \includegraphics[width=\linewidth]{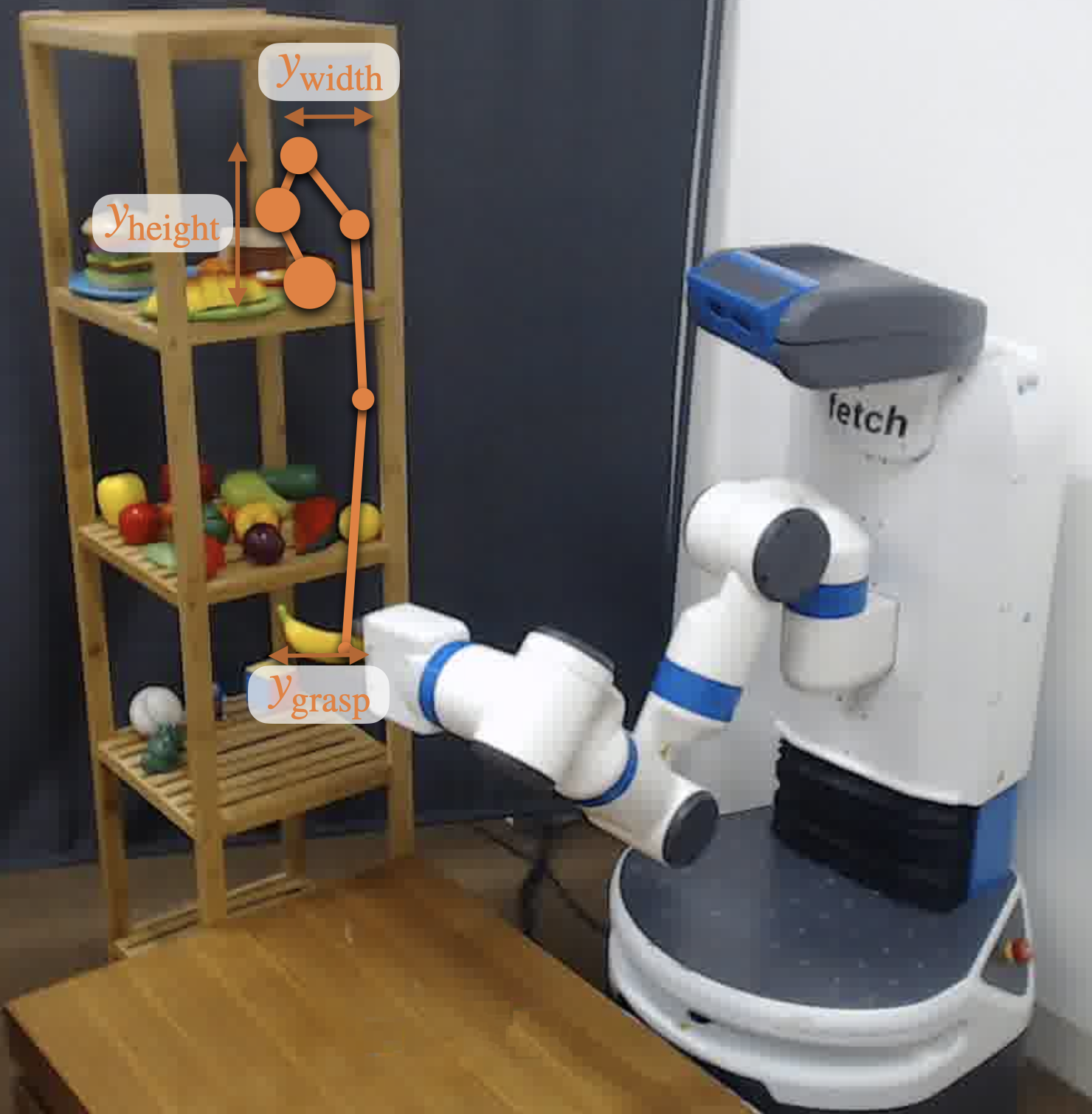}
    \end{minipage}\hskip 2em\begin{minipage}{.45\linewidth}
    \centering
    \begin{tabular}{cr}
    \toprule
         Feature & \multicolumn{1}{c}{Value} \\
     \midrule
        $x_1$ & \popx   \\
        $x_2$ & \popx \\
        $x_3$ & \popx \\
        $y_{\rm speed}$ & \popx\\
                $y_{\rm speed}(1-y_{\rm speed})$ & \popx\\
        $y_{\rm grasp}$ & \popx \\
        $y_{\rm grasp}(1-y_{\rm grasp})$ & \popx \\
        $\htt$ & \popx \\
        $\htt(1-\htt)$ & \popx \\
        $\wtt$ & \popx \\
        $\wtt(1-\wtt)$ & \popx \\
        $1-(y_{\rm grasp} - \wtt)^2$ & \popx\\
        $\succ$ & \popx\\
     \bottomrule
    \end{tabular}
    \end{minipage}
    \caption{Sample Fetch trajectory (left) with extracted features (right).}
    \label{fig:fetchtraj}
\end{figure}

\section{Metrics}\label{app:metrics}
\subsection{MSE}
\label{app:mse}
Our metric is \begin{equation}
    \label{eq:mle}
    \mathcal M_{\text{MSE}} = \sum_{m=1}^M \|\omega_m^* - \widehat\omega_m\|^2_2
\end{equation}
where $\Theta^*=\params[|^*]$ and $\widehat\Theta=\params[\widehat|]$ and the learned reward weights of the experts are matched with the true weights using the Hungarian algorithm. When the learning model assumes a unimodal reward function, as in our simulations for \cref{fig:unimodal_vs_bimodal}, we compute the MSE metric as $\sum_{m=1}^M \|\omega_m^*-\hat\omega\|^2_2$.

\subsection{Log-Likelihood}
\label{app:ll}
Formally, we define the log-likelihood metric as
\begin{equation}
    \label{eq:ll}
    \mathcal M_{\text{LL}} = {\mathbb E}_{{Q\sim\mathcal Q}}\qty[\mathbb E_{x\sim P(X\mid Q)}\log \Pr(x\mid \mathcal Q,D)]
\end{equation} 

for $\mathcal Q$ the uniform distribution across all possible queries and $P(x\mid Q)$ the distribution over the human's response to query $Q$ (as in Eq.~\eqref{eq:pmf}). We can compute the inner term 
\begin{align*}
    \Pr(x\mid Q,\D) &= \mathbb{E}_{\Theta\mid \D} [\Pr(x\mid Q,\Theta)]
\end{align*}

using Metropolis-Hastings as in \cref{sec:overall} to sample from the posterior $\Pr(\Theta\mid \D)$ and computing the inner term with \cref{eq:pmf}.

\subsection{Learned Policy Rewards}
\label{app:learned_policy_rewards}
Similar to the MSE metric, we match the rewards learned via DQN \cite{mnih2013playing} with the true rewards using the Hungarian algorithm.

\section{Experimental Setup}
\label{app:expset}

\subsection{Shelf Descriptions}
\label{app:desc}
A picture of each shelf accompanied the following descriptions. 
\begin{itemize}[nosep]
    \item The top shelf has some space, but you usually put cooked meals there.
    \item The middle shelf is for fruits, but it is already full. The robot may accidentally drop other fruits.
    \item The bottom shelf has a lot of space, but you have been using it for toys.
\end{itemize}

\subsection{User Interface}
\label{app:ui}
\begin{figure*}[t]
    \centering
    \includegraphics[width=\textwidth]{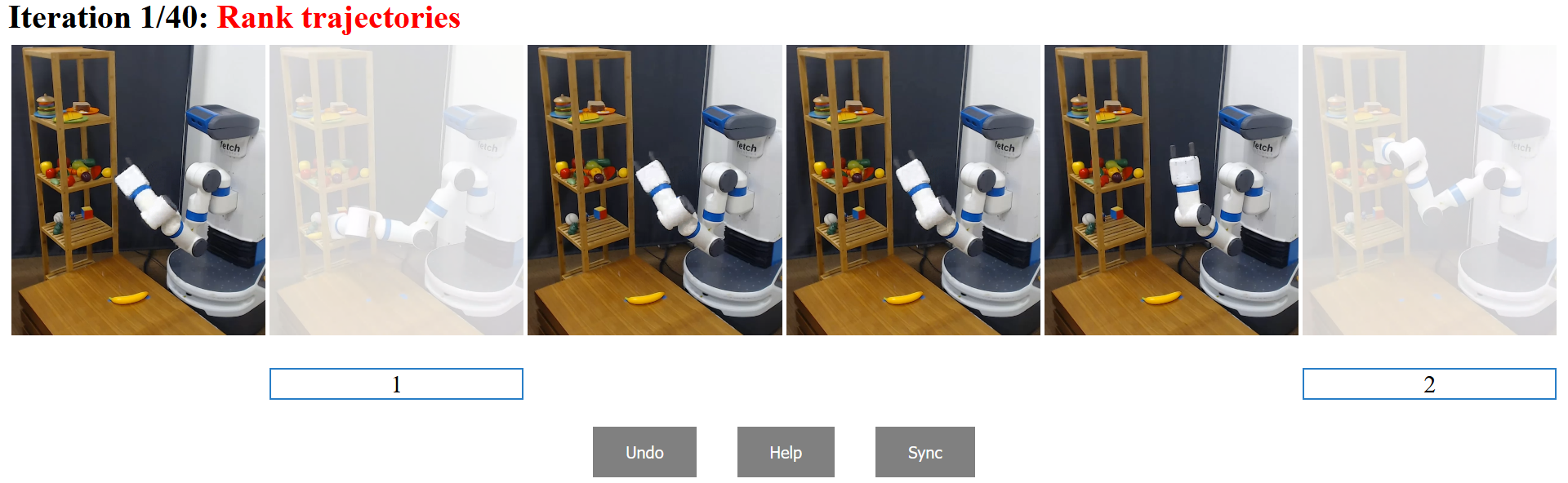}
    \vspace*{-8px}
    \caption{The user interface for the online studies with the real Fetch robot. The user selected the $2^{\textrm{nd}}$ trajectory as their top choice and the $6^{\textrm{th}}$ trajectory as the second top.}
    \label{fig:fetch_ui}
        \vspace*{-23px}
\end{figure*}

For both environments, subjects were told they need to rank the six trajectories in each query by clicking on the trajectories starting from the most preferred to the least. The web interface (see \cref{fig:fetch_ui}) equipped them with ``Undo" and ``Sync" buttons. ``Undo" allowed the subjects to undo a selection they make within a query. ``Sync" enabled them to restart all videos in the query.

\section{Synthetic Experiment}
\label{app:syn}

\subsection{Testing $M>2$}

\begin{wrapfigure}{R}{0.5\linewidth}
    \centering
    \includegraphics[width=\linewidth]{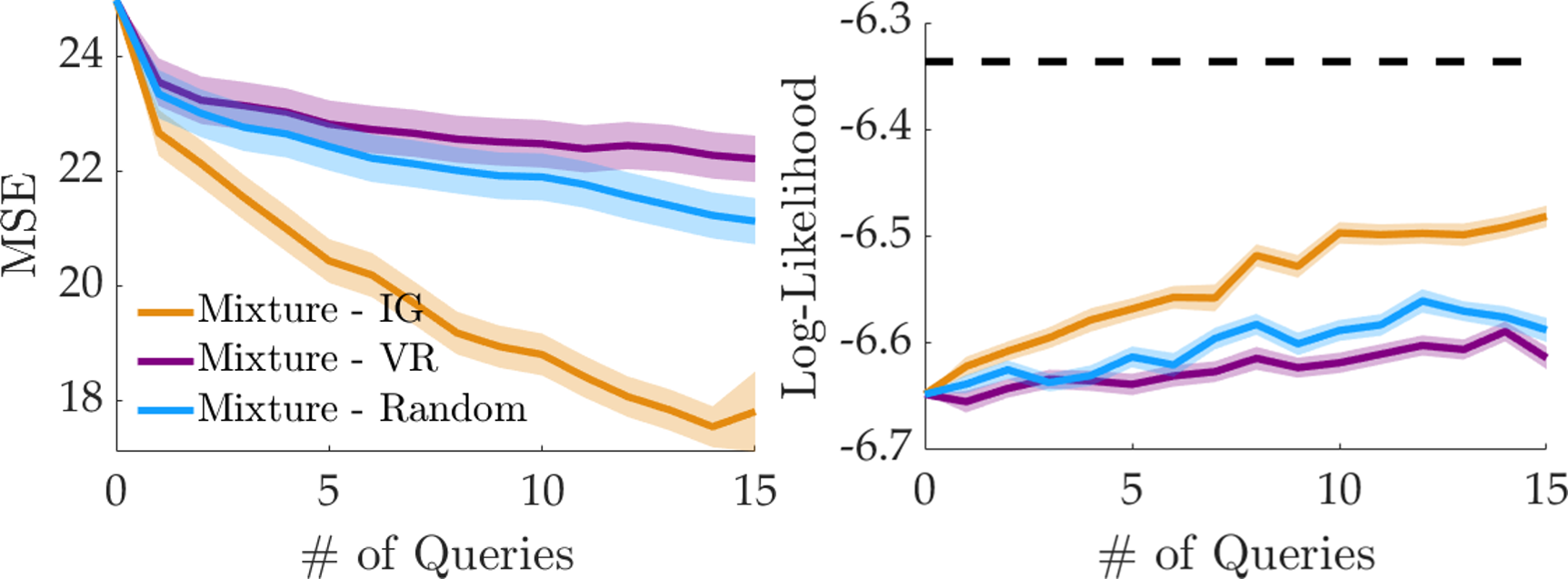}
    \caption{Different querying methods are compared on a synthetic environment (mean$\pm$se over 250 runs).}
    \vspace{-10px}
    \label{fig:synthetic}
\end{wrapfigure}

For our first experiment with synthetic data, we demonstrate effectiveness of our approach for learning mixtures of more than two Plackett-Luce models. In particular, we evaluate our approaches using 250 sets of five randomly simulated reward weights ($M=5$, $K=6$), and trajectory features defined by \begin{revision} $\Phi(\xi_{1:10})\sim \mathcal N(0,I)$, $\Phi(\xi_{11:110})\sim \mathcal N(0,0.1I)$, and $\Phi(\xi_{111:1110})\sim \mathcal N(0,0.01I)$ where 
$I$ is the $3\times3$ identity matrix and $\xi_{n:k}$ refers to the $n$th through $k$th trajectory in the environment
\end{revision}. This environment models complex multimodal structure in the trajectory feature space, which is common to many robotic settings.

\begin{wrapfigure}[15]{R}{0.35\linewidth}
    \centering
    \vspace{-15px}
    \includegraphics[width=\linewidth]{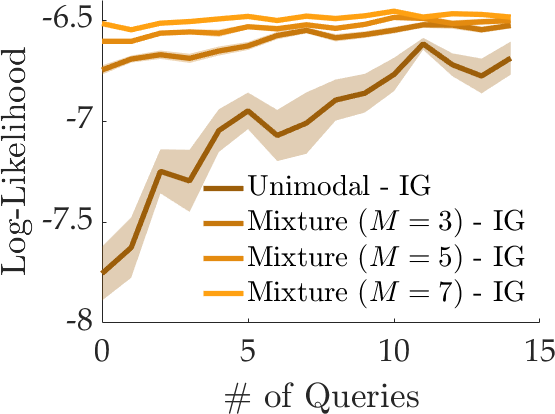}
    \ifrevision \captionsetup{labelfont={color=blue},font={small,color=blue}} \fi
    \caption{
    Different values of $M$ for the IG approach are compared (mean$\pm$se over 100 runs).}
    \label{fig:vary_M}
\end{wrapfigure}

\Cref{fig:synthetic} shows the results of our experiments. We see our approach, Mixture - IG dramatically outperforms the other approaches in both the MSE and log-likelihood metrics.

\begin{revision}
\subsection{Testing Robustness to $M$ Parameter}
\label{app:robust}

We also test the robustness of our Mixture - IG approach to misspecified $M$ values. We repeat the previous experiment, testing the Mixture - IG approach varying the misspecified value of $M$ between $M=1$ (Unimodal) and $M=3,5,7$ (see \Cref{fig:vary_M}). We use the log-likelihood metric since MSE is not well-defined for methods with $M\neq 5$ because they learn a mixture of a different number of reward functions from the true synthetic mixture.

We see the best performance occurs for $M=5$ and $M=7$, with only $M=1$ performing significantly worse. We conclude that in this experiment our Mixture - IG approach is relatively robust to the value of $M$, as long as a sufficiently large value $>1$ is selected.
\end{revision}

\begin{revision}
\section{Additional Unimodal Baseline}
\label{app:new_baseline}
We test an additional baseline on the random queries made during user studies to show the superiority of our learning approach. The additional baseline represents selecting the unimodal reward with fixed norm that maximizes the reward of the top trajectory of each expert-ranked query. We compare this baseline against a learning method that computes the bimodal MLE of the reward function. Formally, for query responses $\D=\bigl\{Q^{(t')},x^{(t')}\bigr\}_{t'=1}^{t}$ with $\xi^{(t')}$ the top trajectory in the ranking $x^{(t')} = (\xi^{(t')}, \ldots)$, we define this baseline to learn the parameters $\theta = \qty{\alpha, \omega}$ where $\alpha=1$ and \begin{align*}
    \tilde \omega &= \sum_{t'=1}^t \Phi(\xi^{(t')})\\
    \omega &= \frac{\tilde\omega}{\|\tilde\omega\|_2}.
\end{align*}

Note that we do not vary the querying method in this experiment. Rather, we compare two methods of learning reward weights from the 15 random human queries that were performed by the Mixture - Random algorithm on the Fetch Robot and LunarLander during our user studies, and then evaluate these methods on the 10 random evaluation queries presented to the humans at the end of the experiment. We compare the two methods in terms of the log-likelihood metric. The results are presented in \Cref{tab:additional}, with our method denoted as ``Mixture MLE" and the new baseline described above denoted as ``Baseline".

\begin{table}[htbp]
\caption{Additional User Study Reward Learning Baseline}
\vskip 1ex
\centering
\label{tab:additional}
\begin{tabular}{rcc}
& LunarLander & Fetch Robot\\
\cmidrule[\heavyrulewidth]{2-3}
\begin{tabular}{r}
\\
\arrayrulecolor{white}\midrule
log-likelihood \\
\\[-5px]
$p$-value
\end{tabular}
&
\begin{tabular}{cc}
\multicolumn{1}{c}{Baseline} & \multicolumn{1}{c}{Mixture MLE}\\\midrule
$-8.23 \pm 0.31$ & $-5.91 \pm 0.18$ \\
& \\[-5px]
\multicolumn{2}{c}{$6.2\cdot 10^{-7}$}
\end{tabular}
&
\begin{tabular}{cc}
\multicolumn{1}{c}{Baseline} & \multicolumn{1}{c}{Mixture MLE}\\
\midrule
$-5.21 \pm 0.22$ & $-4.70 \pm 0.35$\\
& \\[-5px]
\multicolumn{2}{c}{$0.11$}
\end{tabular}
\\
\cmidrule[\heavyrulewidth]{2-3}
\end{tabular}
\end{table}

We see the Mixture MLE method outperforms the Baseline method on both environments, with statistical significance ($p<0.05$) in LunarLander when conducting paired $t$-TESTS.
\end{revision}

\end{document}